\pdfoutput=1
\documentclass[11pt]{article}

\usepackage[preprint]{acl}

\usepackage{times}
\usepackage{latexsym}
\usepackage{xspace}
\usepackage[T1]{fontenc}
\usepackage[utf8]{inputenc}

\usepackage{microtype}

\usepackage{inconsolata}

\usepackage{graphicx}
\usepackage{multirow}
\usepackage{booktabs}
\usepackage{tabularx}
\newcommand{\projname}{SLIDE\xspace}
\newcommand{\projnameSpelledOut}{Sliding Localized Information for Document Extraction}
\title{\projname: \projnameSpelledOut}

\author{Divyansh Singh \\
  University of Florida \\
  \texttt{divyansh.singh@ufl.edu} \\\And
  Manuel Nunez Martinez \\
  University of Florida\\
  \texttt{manuel.nunez@ufl.edu}  \\\AND
  Sonja Schmer Galunder \\
  University of Florida\\
  \texttt{s.schmergalunder@ufl.edu}  \\\And
  Bonnie J. Dorr \\
  University of Florida\\
  \texttt{bonniejdorr@ufl.edu} \\
  }

\begin{document}
\maketitle
\begin{abstract}
Constructing accurate knowledge graphs from long texts and low-resource languages is challenging, as large language models (LLMs) experience degraded performance with longer input chunks. This problem is amplified in low-resource settings where data scarcity hinders accurate entity and relationship extraction. Contextual retrieval methods, while improving retrieval accuracy, struggle with long documents. They truncate critical information in texts exceeding maximum context lengths of LLMs, significantly limiting knowledge graph construction.
We introduce \projname(\projnameSpelledOut), a chunking method that processes long documents by generating local context through overlapping windows. \projname ensures that essential contextual information is retained, enhancing knowledge graph extraction from documents exceeding LLM context limits.
It significantly improves GraphRAG performance, achieving a 24\% increase in entity extraction and a 39\% improvement in relationship extraction for English. For Afrikaans, a low-resource language, \projname achieves a 49\% increase in entity extraction and an 82\% improvement in relationship extraction. Furthermore, it improves upon state-of-the-art in question-answering metrics such as comprehensiveness, diversity and empowerment, demonstrating its effectiveness in multilingual and resource-constrained settings.
\end{abstract}

\section{Introduction}
\begin{figure}[t]
\centering
\includegraphics[width=\columnwidth]{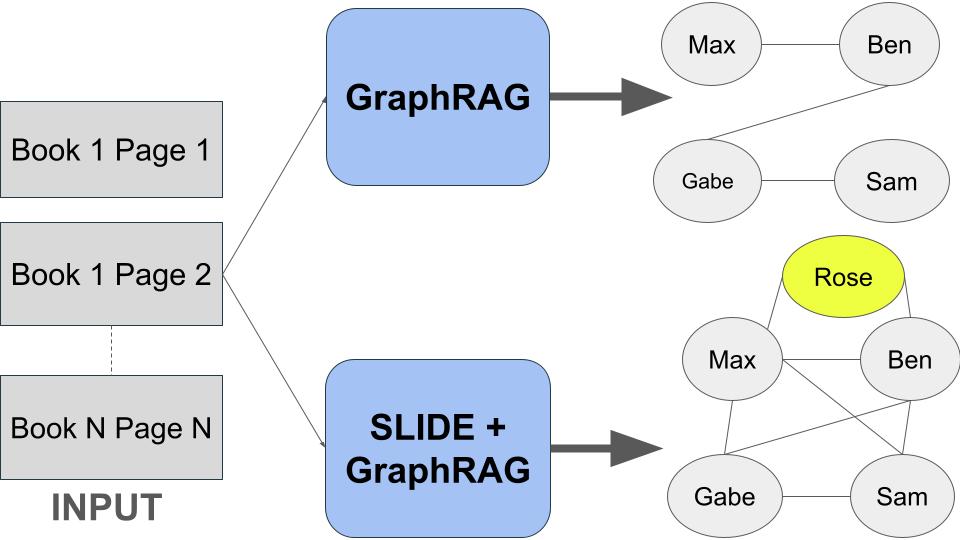} 
\vspace*{-.2in}
\caption{Comparison of knowledge graph extraction from a set a of fiction book pages as chunks (Top) Using GraphRAG without \projname produces a knowledge graph with fewer nodes (representing entities) and edges (representing relationships). (Bottom) Using \projname results in a richer knowledge graph.}
\vspace*{-.2in}
\label{fig:comparison_chunking}
\end{figure}
In the past year, integrating knowledge graphs with large language models (LLMs) has enhanced reasoning, summarization, and retrieval-augmented generation (RAG) \cite{graphrag,lightrag,chang2024communitykgragleveragingcommunitystructures}. GraphRAG, a retrieval-augmented generation framework, uses LLMs to extract knowledge graphs for query-focused summarization \cite{graphrag}.

However, Constructing accurate knowledge graphs from long documents and datasets in low-resource languages presents significant challenges when using large language models (LLMs). Current methods struggle to extract nuanced entities and relationships, reducing their effectiveness in multilingual and resource-constrained settings.

Large language models (LLMs), such as GPT-4o \cite{achiam2023gpt}, exhibit diminished effectiveness in entity and relationship extraction as input chunk lengths increase, leading to degraded accuracy for longer texts \cite{graphrag}. Additionally, LLMs struggle with relationship extraction in low-resource languages \cite{chen2024translationfusionimproveszeroshot,jinensibieke2024goodllmsrelationextraction} due to data scarcity, further limiting their applicability in constructing knowledge graphs in diverse linguistic contexts.

Contextual retrieval \cite{contextual} involves embedding each chunk of text along with its context from the entire document to enhance retrieval accuracy; it faces significant limitations with longer documents. Embedding each chunk with full document context becomes computationally expensive and inefficient, particularly when the document length exceeds the maximum context length of LLMs, leading to truncated or omitted critical information. \cite{jiang2024longllmlinguaacceleratingenhancingllms,li2024longcontextllmsstrugglelong}.

To tackle these challenges, we introduce \projname (\projnameSpelledOut), a chunking method designed to process documents that exceed the token limits of large language models (LLMs). SLIDE generates local context through overlapping windows, allowing efficient handling of long documents while maintaining contextual integrity. Unlike contextual retrieval, \projname limits the maximum number of tokens to the size of the window used, optimizing the handling of longer inputs.
Furthermore, we apply this method and demonstrate improved knowledge graph extraction (See Figure \ref{fig:comparison_chunking}) for GraphRAG systems in both English and Afrikaans, a low-resource language spoken by 7 million people in South Africa and Namibia \cite{afrikaans}, demonstrating its effectiveness in multilingual contexts and its potential to enhance knowledge extraction in resource-constrained settings.

This paper makes the following key contributions:
(1) Introduction of a sliding window-based contextual chunking method (\projname) for enhanced knowledge graph construction. (2) Ability to handle documents longer than the context length of the LLM by limiting token usage to the size of the sliding window, in contrast to the full-document context used in Contextual Retrieval. (3) Demonstrate evidence for performance increment over existing GraphRAG systems in both English and Afrikaans.
Specifically, SLIDE achieved a 24\% higher entity extraction and a 39\% higher relationship extraction for English, while for the low-resource language Afrikaans, it saw a 49\% increase in entity extraction and an 82\% increase in relationship extraction. Additionally, SLIDE outperformed the state of the art on key question-answering metrics, including Comprehensiveness, Diversity and Empowerment.

\section{Related Work}
Since \projname enhances knowledge graph construction in GraphRAG systems through contextual chunking, we first discuss related work in GraphRAG and chunking, highlighting their strengths and limitations. This sets the stage for our approach, which builds on GraphRAG by using overlapping windows to improve entity and relationship extraction.
\subsection{GraphRAG and Knowledge Graphs}
GraphRAG \cite{graphrag} is an advanced RAG framework that integrates knowledge graphs with large language models (LLMs) \cite{trajanoska2023enhancingknowledgegraphconstruction} to enhance reasoning and contextual understanding. Unlike traditional RAG systems, GraphRAG builds a knowledge graph with entities as nodes and relationships as edges, enabling precise and context-rich responses by leveraging the graph's structure \cite{graphrag,wu2024medicalgraphragsafe}. Large language models (LLMs), such as GPT-4, show reduced effectiveness in entity and relationship extraction as input chunk lengths increase, degrading accuracy for longer texts \cite{graphrag}. They also struggle with relationship extraction in low-resource languages, limiting their applicability \cite{chen2024translationfusionimproveszeroshot, jinensibieke2024goodllmsrelationextraction}.
Building upon this work, our approach further enhances knowledge graph extraction by incorporating localized context which improves entity and relationship extraction.
\subsection{Contextual Chunking}
Recent work in RAG systems has explored advanced chunking techniques to enhance retrieval and knowledge graph construction. \citet{gunther2024late} implemented late chunking, where entire documents are embedded to capture global context before splitting into chunks, improving retrieval by emphasizing document-level coherence. However, this focus on global embeddings is less suited for knowledge graph construction. Our method instead uses localized context from raw text to retain meaningful relationships for improved entity and relationship extraction.

\citet{wu2024medicalgraphragsafe} introduced a hybrid chunking approach for Medical Graph RAG, combining structural markers like paragraphs with semantic coherence to produce self-contained chunks. While effective, this approach relies on predefined boundaries. Our method extends this by generating contextual information from neighboring chunks, enhancing the completeness of knowledge graph construction.

Contextual retrieval \cite{contextual} improves accuracy but struggles with longer documents, as embedding each chunk with full document context is computationally expensive and truncates critical information with documents exceeding maximum context length of the model \cite{jiang2024longllmlinguaacceleratingenhancingllms,li2024longcontextllmsstrugglelong}. Our overlapping window-based approach addresses these inefficiencies, improving performance in both retrieval and knowledge graph construction.

\begin{figure}[t]
\centering
\includegraphics[width=\columnwidth]{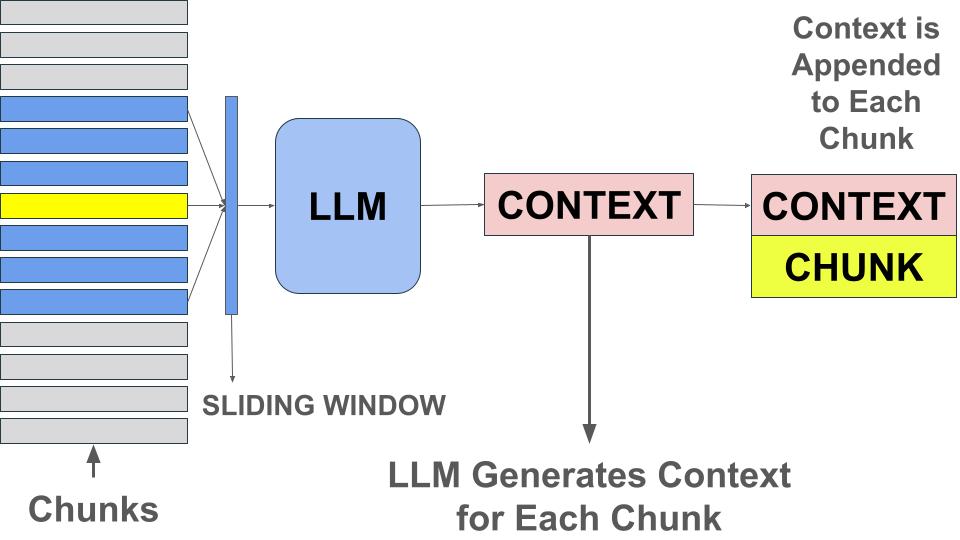} 
\vspace*{-.2in}
\caption{How Context Is Generated for Each Chunk Using a Sliding Window Approach to Form a \textit{Contextual Chunk}. Each Chunk, Along with Its Neighboring Chunks, Is Fed to an LLM to Generate Context.}
\vspace*{-.2in}
\label{fig:context_intro}
\end{figure}
\section{\projname : \projnameSpelledOut}
\label{sec:contextual_chunking}

Given a document \( D \) and chunks \( C_1, C_2, \dots, C_k \) based on the desired token length (similar to \citet{graphrag}), \projname employs a \textit{sliding window} technique to incorporate neighboring chunks within a fixed window size. The window size \(n\), a hyperparameter, determined by the LLM's context length and computational constraints, is set to 10. For each chunk \( C_i \), a window of \( \frac{n}{2} \) preceding and \( \frac{n}{2} \) succeeding chunks is included, resulting in a total of \( n+1 \) chunks being fed to the LLM. This ensures each chunk is enriched with meaningful context from its surroundings (see Figure \ref{fig:context_intro}). Further details on the prompt design are provided in Appendix \ref{sec:appendix-a}. The LLM generates a concise context in English that situates \( C_i \) within its neighborhood, and this process is repeated for all chunks, appending the generated context to each original chunk.

\subsection{Knowledge Graph Construction}

Once each chunk is appended with its generated context, chunks generated by \projname are processed for knowledge graph construction following the approach outlined in \cite{graphrag}. The additional context enhances entity and relationship extraction by capturing a greater number of entities and relationships (see Table \ref{tab:entities}) suggesting a better entity resolution due to additional context and generation of more informative community summaries and claims. 

\begin{table}[h!]
\centering
\small % Reduce font size
\begin{tabular}{lcccc}
\toprule
\textbf{Chunk Size} & \textbf{Gleanings} & \textbf{\projname} & \textbf{Without \projname} \\
\midrule
\multirow{2}{*}{1200} & 1 & 341, 784 & 325, 613 \\
                      & 4 & 350, 800 & 345, 630 \\
\midrule
\multirow{2}{*}{2400} & 1 & 235, 488 & 169, 385 \\
                      & 4 & 239, 495 & 172, 398 \\
\bottomrule
\end{tabular}
\caption{Comparison of Number of Entities and Relationships (values are comma-separated) Extracted for English, With and Without \projname}
\label{tab:entities}
\end{table}

\begin{table}[h!]
\centering
\small % Reduce font size
\begin{tabular}{lcc}
\toprule
\textbf{Gleanings} & \textbf{\projname} & \textbf{Without \projname} \\
\midrule
1  & 472, 758 & 344, 433 \\
4  & 512, 792 & 348, 435 \\
\bottomrule
\end{tabular}
\caption{Comparison of Number of Entities and Relationships (values are comma-separated) Extracted for Afrikaans, With and Without \projname (Chunk Size: 1200)}
\label{tab:entities_afrikaans}
\end{table}

\subsection{Enhancing Knowledge Graphs in Afrikaans: A low resource language}

We further demonstrate that adding English context to situate a chunk within its neighborhood also significantly enhances entity and relationship coverage in Afrikaans, a low-resource language. By leveraging the richer context from English, the LLM captures substantially more entities and relationships in Afrikaans (See Table \ref{tab:entities_afrikaans}), thereby driving a more robust knowledge graph construction for multilingual and low-resource scenarios.

\begin{table*}[t]
\small
\begin{tabular}{|l|l|l|l|l|l|l|l|l|l|l|l|}
\hline
 & \multicolumn{2}{|c|}{Level 1} & \multicolumn{2}{|c|}{Level 2} & \multicolumn{2}{|c|}{Level 3} & \multicolumn{2}{|c|}{Overall Count} \\\hline
 & English & Afrikaans & English & Afrikaans & English & Afrikaans & English & Afrikaans \\\hline
Comprehensiveness & \textbf{69}/22 & \textbf{87}/8 & \textbf{71}/21 & \textbf{91}/5 & \textbf{76}/15 & \textbf{84}/12 & \textbf{216}/58 & \textbf{262}/25 \\\hline
Diversity         & \textbf{70}/21 & \textbf{85}/10 & \textbf{70}/22 & \textbf{87}/9 & \textbf{76}/15 & \textbf{80}/16 & \textbf{216}/58 & \textbf{252}/35 \\\hline
Empowerment       & \textbf{70}/21 & \textbf{82}/13 & \textbf{70}/22 & \textbf{86}/10 & \textbf{73}/18 & \textbf{80}/16 & \textbf{213}/61 & \textbf{248}/39 \\\hline
\end{tabular}
\caption{Comparison of Winning Response to the 100 Questions for each community level: With \projname vs Without \projname for English and Afrikaans. Values in each cell are separated by /. \textit{Ties are not indicated in the table.}}
\label{tab:comparison_chunks_merged}
\end{table*}
\section{Experimental Setup}
This section outlines the experimental setup we use to evaluate the improvement of \projname in the performance of GraphRAG. Our computations are using 4 cores of an AMD 32-Core Processor and 3 NVIDIA A100 GPUs.
We conduct an ablation study to analyze how \projname improves GraphRAG.

\subsection{Datasets}

\begin{itemize}
        \item Pride and Prejudice by Jane Austen (English):\footnote{\url{https://www.gutenberg.org/ebooks/1342}}
        This fiction book contains over 160,000 tokens, making it an exemplary case of a long text document. Its intricate narrative and complex character dynamics provide a rich dataset for excracting and building knowledge graphs.
        \item Bang vir die lewe by Henry Bordeaux (Afrikaans):\footnote{\url{https://www.gutenberg.org/ebooks/72345}}
        It is an Afrikaans novel containing approximately 150,000 tokens. This work exemplifies a long text document in a low-resource language.
\end{itemize}
\subsection{Contextual Chunks and Knowledge Graph Extraction}

We evaluate the impact of \projname on knowledge graph extraction using a chunk size of 1200 tokens. The sliding window size was set to include 11 chunks (5 chunks before and after the reference chunk) without overlap, enabling an independent assessment.

We use the \texttt{Hermes 3 LLAMA 3.1 70B-Instruct} \cite{llama,teknium2024hermes3technicalreport} as the LLM for our experiments and the \texttt{E5-mistral-7b-instruct} \cite{wang2023improving} model to produce embeddings. The GraphRAG module\footnote{\url{https://microsoft.github.io/graphrag/}} was used to extract and query knowledge graphs, with 4 gleanings.

\subsection{Evaluation Metrics}

LLMs have been shown to achieve state-of-the-art or competitive results compared to human judgments \cite{wang2023chatgpt,zheng2023judging}. Thus we adopt the evaluation methodology similar to the one used by \citet{graphrag} and \citet{lightrag}. Leveraging LLMs as evaluators involves assessing the quality of generated outputs through head-to-head comparisons of competing outputs.

For each language, we generate 100 questions using the RAGAS \cite{es2023ragasautomatedevaluationretrieval}. This approach ensured a diverse set of questions that varied in complexity and context requirements, providing a robust evaluation of the model's performance.

For our evaluation, we use the following metrics:
\textbf{Comprehensiveness}: Measures how much detail the answer provides to cover all aspects of the question. \textbf{Diversity}: Assesses how varied and rich the answer is in providing different perspectives and insights. \textbf{Empowerment}: Evaluates how well the answer helps the reader understand and make informed judgments about the topic.

The LLM is provided with a question, target metric, and a pair of answers. It evaluates which answer performs better or declares a tie if differences are negligible and provides reasoning for its evaluation. Each comparison is repeated five times to account and mean scores are computed for final evaluation. According to \citet{graphrag}, the best community levels to query are 1, 2, and 3. We query all of these levels for a comprehensive evaluation.

\section{Results}

Our evaluation (See Table \ref{tab:comparison_chunks_merged}) for English and Afrikaans demonstrates chunks  generated using \projname consistently outperform standard chunks across most metrics.

For English, chunks generated by \projname significantly lead in Comprehensiveness, Diversity, and Empowerment by 53\%, 53\%, and 51\%, respectively, compared to standard chunks.

In Afrikaans, \projname chunks show substantial improvements in all metrics, leading by 79\%, 72\%, and 70\% in Comprehensiveness, Diversity, and Empowerment, respectively.

These results indicate that \projname produces more informative community summaries and claims for GraphRAG, which improve the quality of responses to queries.

Additionally, entity and relationship counts (See tables \ref{tab:entities} and \ref{tab:entities_afrikaans}) highlight an enhanced coverage of the knowledge graphs when using \projname, further supporting these performance enhancements.

\section{Conclusion}

Overall, \projname achieves improved performance across evaluation metrics for both English and Afrikaans by providing richer and more detailed responses. \projname also demonstrates an increased entity and relationship coverage compared to standard chunking. These findings validate the advantage of using \projname for enhanced knowledge graph extraction and suggests potential for language-specific optimizations.

\section*{Limitations}
Despite its strengths, \projname has few limitations. The sliding window approach employed by SLIDE significantly increases computational overhead. For each chunk $C_i$, the inclusion of neighboring chunks within a fixed window size results in repetitive processing of overlapping text. This redundancy amplifies the computational cost, particularly for large documents, as the same chunks are processed multiple times. Such inefficiency can be a bottleneck when scaling to extensive datasets or deploying on computational resource-constrained systems.

\projname's performance is also limited with the capabilities of the underlying LLM. Using advanced models such as GPT-4o could potentially yield better results due to their superior contextual understanding and reasoning abilities. However, computational constraints often necessitate reliance on less powerful models, which can limit the quality of extracted knowledge graphs.

Additionally, while \projname demonstrates increased entity and relationship coverage, it does not evaluate the accuracy of the extracted entities and relationships using standard Named Entity Recognition (NER) metrics such as precision and recall. Without such evaluation, it remains unclear whether the improvements in coverage come at the cost of increased false positives or reduced extraction reliability.
Finally, the evaluation methodology for SLIDE relies on LLAMA 3.1 as a judge for assessing entity and relationship extraction and generating questions using RAGAS. LLAMA 3.1 itself may not be an optimal evaluator due to potential biases \cite{wataoka2024self} and limitations in its understanding of low-resource languages like Afrikaans \cite{Alam2024LLMsFL}. This raises concerns about the reliability and generalizability of the reported improvements.

These limitations highlight important areas for improvement and will be the focus of future work. Addressing them would require innovations in computational efficiency, advancements in multilingual LLMs, and more robust evaluation frameworks to ensure consistent and scalable performance across diverse use cases.
\section*{Ethics Statement}
Our work complies with the policies of the Meta LLaMA Terms and Conditions. No personal or sensitive information has been collected. We exclusively utilize publicly available datasets and examine these without intervening or interacting with any individuals.

\section*{Acknowledgement}

This work is supported, in part, by DARPA. Any opinions, findings and conclusions or recommendations expressed in this material are those of the authors and do not necessarily reflect the views of the US Government.
\bibliography{custom}

\appendix

\section{Prompt for Context Generation}
\label{sec:appendix-a}
System:
\texttt{You are an assistant which generates short English context to situate the input chunks in the input document. Failure to adhere to this guideline will get you terminated.} 

Human:
\texttt{Here is the document:}  
\texttt{\{bigger\_chunk\}}  

\texttt{Here is the chunk we want to situate within the whole document:}  
\texttt{\{chunk\}}  

\texttt{Please give English context to situate this chunk within the overall document for the purposes of improving search retrieval of the chunk.}  
\texttt{Answer only with a short context. Do not provide any additional text.}

\end{document}